\documentclass[10pt,twocolumn,letterpaper]{article}

\usepackage{iccv}
\usepackage{times}
\usepackage{epsfig}
\usepackage{graphicx}
\usepackage{amsmath}
\usepackage{amssymb}
\usepackage{xcolor}
\usepackage{multirow}
\usepackage{multicol}
\usepackage{booktabs}
\usepackage{subcaption}
\usepackage[T1]{fontenc}  
\usepackage[utf8]{inputenc} 
\usepackage{authblk}

\usepackage[breaklinks=true,bookmarks=false]{hyperref}

\iccvfinalcopy 

\def\iccvPaperID{****} 

\ificcvfinal\fi

\begin{document}

\title{Augmented Transformer with Adaptive Graph for\\ Temporal Action Proposal Generation}



\author{Shuning Chang\textsuperscript{\rm 1}\textsuperscript{\rm 2}\thanks{Work done during an internship at Alibaba Group.}, Pichao Wang\textsuperscript{\rm 2}\thanks{Corresponding author.}, Fan Wang\textsuperscript{\rm 2}, Hao Li\textsuperscript{\rm 2}, Jiashi Feng\textsuperscript{\rm 1}, \\ 
\textsuperscript{\rm 1}National University of Singapore \quad \textsuperscript{\rm 2} Alibaba Group\\
{changshuning@u.nus.edu},\,{\ \{pichao.wang,fan.w,lihao.lh\}@alibaba-inc.com},\,{\ elefjia@nus.edu} 
}

\maketitle

\ificcvfinal\thispagestyle{empty}\fi

\begin{abstract}
Temporal action proposal generation (TAPG) is a fundamental and challenging task in video understanding, especially in temporal action detection. Most previous works focus on capturing the local temporal context and can well locate simple action instances with clean frames and clear boundaries. However, they generally fail in complicated scenarios where interested actions involve irrelevant frames and background clutters, and the local temporal context becomes less effective. 
To deal with these problems, we present an augmented transformer with adaptive graph network (ATAG)  to exploit both long-range and local temporal contexts for TAPG. 
Specifically, we enhance the vanilla transformer by equipping a snippet actionness loss and a front block, dubbed augmented transformer, and it improves the abilities of capturing long-range dependencies and learning robust feature for noisy action instances.
Moreover, an adaptive graph convolutional network (GCN) is proposed to build local temporal context by mining the position information and difference between adjacent features.
The features from the two modules carry rich semantic information of the video, and are fused for effective sequential proposal generation. Extensive experiments are conducted on two challenging datasets, THUMOS14 and ActivityNet1.3, and the results demonstrate that our method outperforms state-of-the-art TAPG methods. Our code will be released soon.
\end{abstract}


\section{Introduction}
Temporal action detection, which aims at locating specified actions from untrimmed videos in the temporal dimension, has been widely applied to various tasks such as video understanding~\cite{sun2015temporal,wang2018rgb}, surveillance~\cite{sultani2018real}, \etc.
Similar to object detection, the pipeline of temporal action detection is usually divided into two stages: temporal action proposal generation (TAPG) and action classification.
The performance of such two-stage temporal action detectors is largely determined by the proposal quality from TAPG.

\begin{figure}[t]
\begin{subfigure}{.5\textwidth}
\includegraphics[width=0.9\linewidth]{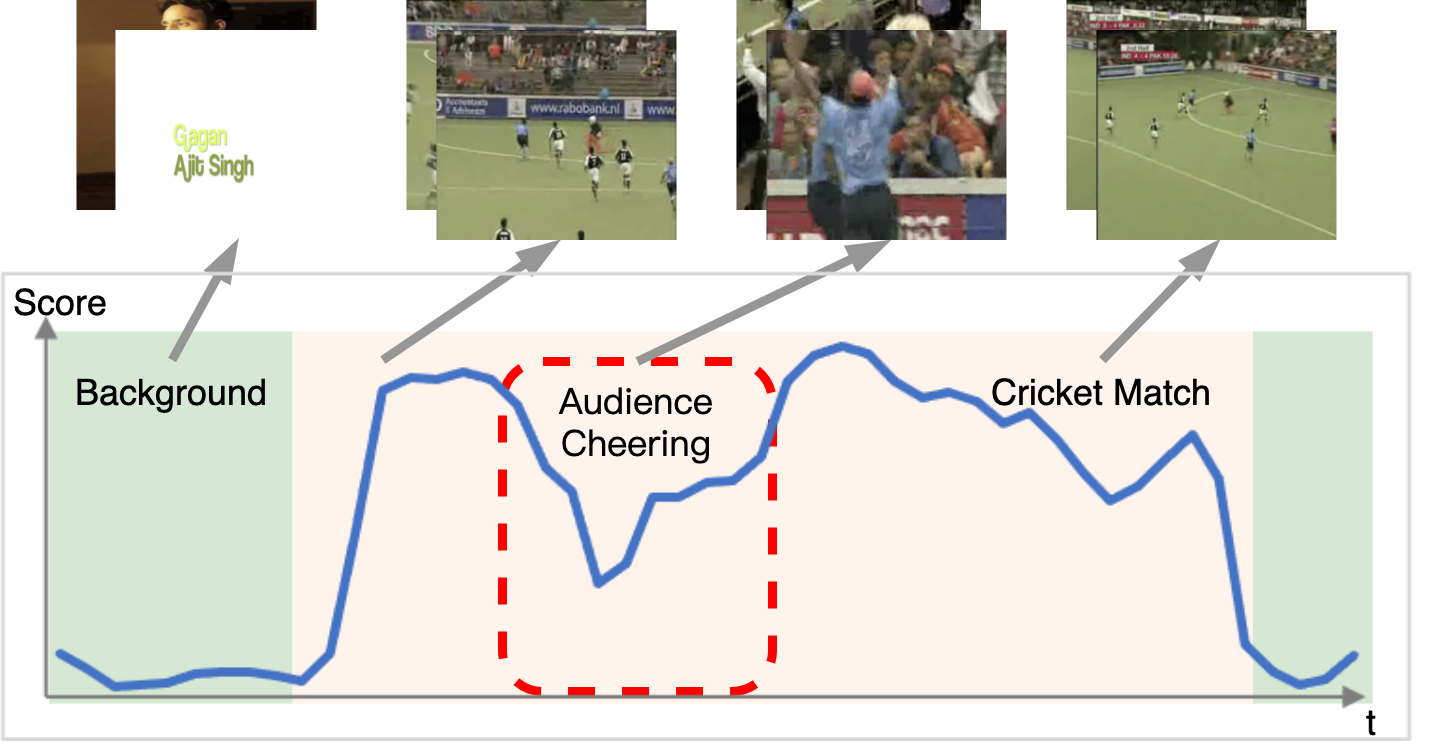}
\centering
\caption{BMN+Actionness predictor. }
\label{figure1:a}
\end{subfigure}
\begin{subfigure}{.5\textwidth}
\includegraphics[width=0.9\linewidth]{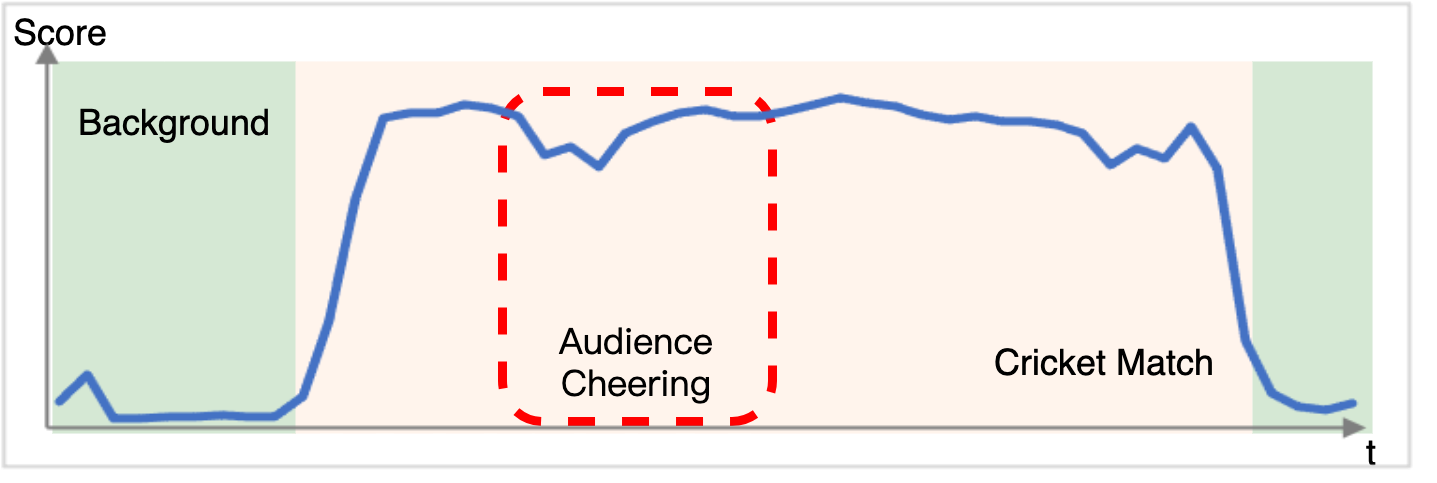}
\centering
\caption{Global context capture. }
\label{figure1:b}
\end{subfigure}
\caption{Benefit of using long-range dependencies. 
We apply BMN~\cite{lin2019bmn} and our method to detect the specified action \textit{cricket match} in a video segment, which also includes an audience cheering segment highlighted in red dotted boxes.
We plot the curves of snippet-wise actionness scores for predictions by BMN without using global context in (a) and our method using global context in (b).
It can be seen the one with global context shows higher confidence scores for the frames of audience cheering, demonstrating the benefits brought by global context for temporal action proposal generation.}
\label{figure1}
\end{figure}

Prior works focus on exploiting the local temporal context information~\cite{chao2018rethinking,lin2019bmn,chao2018rethinking,xu2020g} for generating high quality proposals.
For example,~\cite{chao2018rethinking,lin2019bmn} apply a pre-defined ratio to extend the temporal action boundaries, and use temporal convolutional layers or dilated convolutional layers to increase the receptive field;~\cite{xu2020g} leverages a GCN model to aggregate local temporal and semantic context.
Using local context information can effectively build relations among neighbors of  current snippets, which works well for locating simple action instances with clean frames and clear boundaries, but is sensitive to action/background representation. 
However, the real-world cases are often complicated, with many irrelevant frames and background clutters over the interested actions.
Only exploiting the local temporal context lacks the semantic understanding for the whole video, often leading to failures of locating precise boundaries.
As shown in Figure \ref{figure1:a}, an actionness predictor is added into a recent method BMN~\cite{lin2019bmn} to show its performance on a complicated video.
We can see that, if we only leverage local temporal context for detecting the specified action (i.e. \textit{cricket match}), the long audience cheering segment embedded in frames of the specified action obviously gives low action confidence scores.
But if a longer range is observed (in Figure \ref{figure1:b}), the model is able to understand that the audience cheering is also part of the match, and thus treating this segment and the actions before and after as a whole can avoid wrong boundary prediction.
From this example, 
we claim that long-range temporal dependencies should also be exploited for more comprehensive video understanding in complicated situations.
In this work, we propose an augmented transformer with adaptive graph network (ATAG) to take advantage of the two types of contexts for effective temporal proposal generation. It includes an augmented transformer which 
mines long-range temporal context for noisy action instance localization, and an adaptive GCN which captures local temporal context.


Specifically, 
to locate noisy action instance, one way is to adequately understand the holistic video and regard the noisy frames as a part of action according to the semantics. The transformer is adopted to achieve this purpose, which has shown superior capability of capturing long-term dependencies~\cite{Girdhar_2019_CVPR,han2020mm}.
However, the vanilla transformer has two issues if directly applied in our task. First, the loss functions in the traditional TAPG methods only supervise proposal-level signal and cannot directly guide the transformer to aggregate long-range features to snippet-level features.
Therefore, we apply a snippet actionness loss to binarily classify each snippet into action or background according to the snippet-level features output by the transformer. 
The snippets, especially for those noisy snippets, need to extract helpful information from other snippets to be correctly classified, thus the snippet actionness loss explicitly forces the transformer to selectively learn long-range dependencies.
On the other hand, in the vanilla transformer, when generating the queries and keys, the receptive field is constraint to the snippet itself and it is not a robust feature learning style for these noisy snippets. To deal with this problem, we introduce a front block on top of vanilla transformer. The front block is a convolution-based lightweight network, which expands the temporal receptive field and filters out noisy frame features. 

For the local temporal context capture, we focus on exploiting the position and gradient/difference information between adjacent features.
We propose an adaptive GCN to build the local temporal context, 
where two adjacency matrices are carefully designed. One matrix has all the elements generated during training. It is similar to conventional convolutional layers but different positions correspond to different kernels,  which represents the common pattern for all the training data and position information. It can adaptively determine whether the farther snippets should have smaller weights than the central one, or whether the snippets near the edge of the video should be assigned with smaller weights.
The other is a content-based adjacency matrix, which captures the difference between node pairs and represents the unique pattern for each data.
By combining the above two adjacency matrices, our data-driven graph increases the flexibility for graph construction and brings more generality to build local context relationships in various samples.


We demonstrate the effectiveness of ATAG on two challenging datasets, THUMOS14 and ActivityNet13. Our model outperforms well established  state-of-the-art methods significantly.

\begin{figure*}
\includegraphics[width=1.\linewidth]{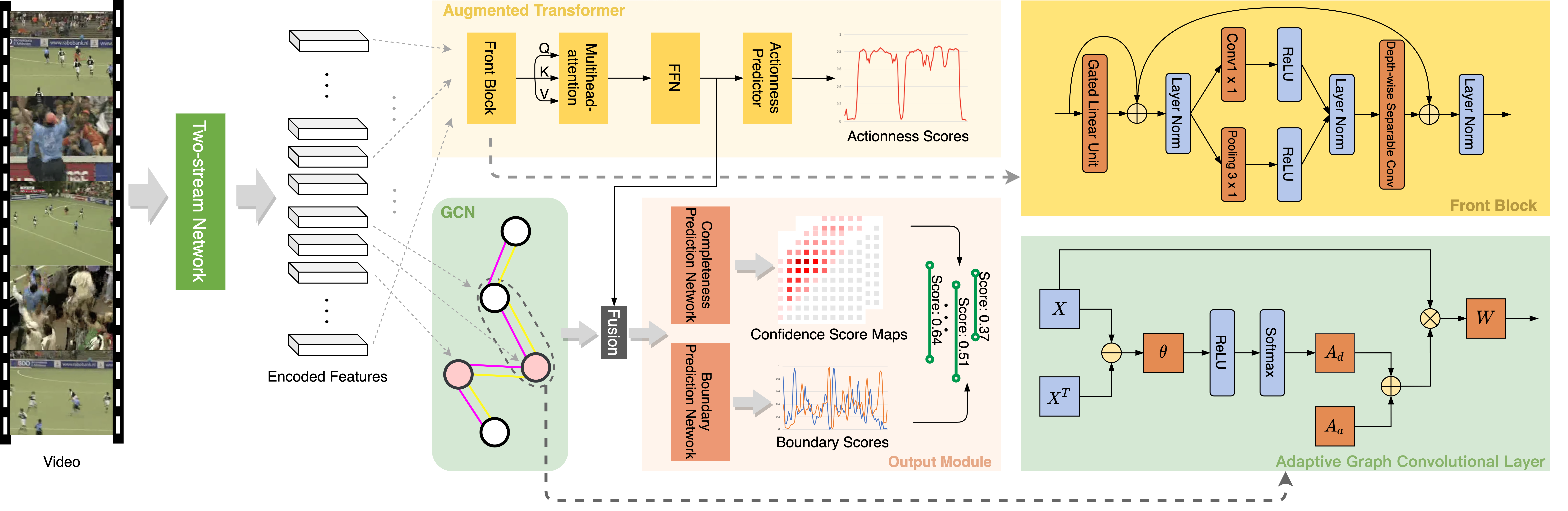}
\centering
\caption{Illustration on the architecture of our method. It first employs a two-stream network to extract snippet-level encoded features. Then, it deploys an augmented transformer and an adaptive graph to capture global and local context, respectively. After fusing global and local features, an output module generates the desired predictions.}
\label{figure:structure}
\end{figure*}

\section{Related work}
\paragraph{Attention mechanism for long-range dependencies.}
Long-range attention mechanism is to compute the response at a position in a sequence by accessing all positions and taking their weighted average in the embedding space.
It is broadly leveraged in natural language processing (NLP) and computer vision.
Vaswani et al.~\cite{vaswani2017attention}
introduce a self-attention mechanism, called transformer, capturing long-range dependencies among words in one sentence to address the machine translation task. 
Following~\cite{vaswani2017attention}, many transformer-based models are proposed and show great potential to tackle various tasks.
Devlin et al.~\cite{devlin2018bert} apply the bidirectional training of Transformer and successfully deal with a broad set of NLP tasks.
Girdhar et al.~\cite{Girdhar_2019_CVPR} utilize transformer to aggregate features from the spatiotemporal context for recognizing and localizing human actions.
Carion et al.~\cite{carion2020endtoend} employ transformers to build an end-to-end object detection model.
Besides transformers, there are some other manners to utilize the attention mechanism to capture long-range dependencies.
For example, Wang et al.~\cite{Wang_2018_CVPR} embed non-local structure into the action recognition network to capture long-range dependencies.
Wu et al.~\cite{Wu_2019_CVPR} introduce long-term feature banks to analyze videos.
How to effectively exploit temporal context is a pivotal problem in video analysis and long-range dependencies has been applied widely~\cite{Girdhar_2019_CVPR,Wang_2018_CVPR,Wu_2019_CVPR,kozlov2020lightweight}. 
However, the long-range dependencies for TAPG has not been well explored, especially in the complicated noisy scenarios.


\paragraph{Temporal action proposal generation (TAPG).}
Early TAPG methods use temporal sliding window~\cite{oneata2014lear,wang2014action} and snippet-wise probability~\cite{shou2016temporal,zhao2017temporal} to generate candidate proposals. 
The former fails to handle ground truth action instances with various durations, while the latter is sensitive to noise and can hardly handle actions with long duration.
Recently, most methods adopt multi-scale anchors to generate proposals~\cite{chao2018rethinking,lin2018bsn,Liu_2019_CVPR,lin2019bmn,xu2020g,zhao2020bottom,bai2020boundary}, similar with the idea in anchor-based object detection~\cite{ren2015faster}.
Chao et al.~\cite{chao2018rethinking} adopt a multi-tower architecture to align the receptive field and extend proposal boundaries by a pre-defined ratio which is widely applied by later methods.
In BSN~\cite{lin2018bsn}, Lin et al. directly predict the probability of boundaries and cover flexible temporal duration of action instances. 
Based on BSN, Lin et al.~\cite{lin2019bmn} further propose a boundary-matching mechanism to evaluate the completeness of densely distributed proposals.
The ideas of GCN have been adopted in some work,
Xu et al.~\cite{xu2020g} extract temporal context and semantic context by GCN, but applies conventional temporal convolutional layers for actual implementation in temporal context extraction.
Bai et al.~\cite{bai2020boundary} perform a GCN to build the  relationship between the boundaries and the content. 
Therefore, they do not apply GCN to capture local temporal context.

Above methods have made significant progress in this field. However, they have limited capability of building local dependencies from the local context, leading to degraded performance for complicated cases. 
Our proposed ATAG has the ability to build both long-range and local temporal dependencies to achieve better video understanding.

\section{Method}
The overall architecture of the proposed augmented transformer with adaptive graph network (ATAG) is shown in Fig.~\ref{figure:structure}.
In this section, we  first describe the problem formulation and backbone feature extraction in Sec.~\ref{ssec:formulation}.
The core augmented transformer with adaptive graph architecture is then introduced in Sec.~\ref{ssec:transformer}, including augmented transformer and adaptive GCN as well, followed by the output module (Sec.~\ref{ssec:output}) used to generate final prediction, and the training~(Sec.~\ref{ssec:training}) and inference~(Sec.~\ref{ssec:inference}) stage. 

\subsection{Problem formulation}
\label{ssec:formulation}
We denote an untrimmed video $\mathcal{X}$ as a frame sequence $\mathcal{X} = \{x_n\}^{l}_{n=1}$ with $l$ frames, where $x_n$ is its 
$n$-th RGB frame; and  the temporal annotations for the $N_g$  action instances 
included in $\mathcal{X}$ are denoted as $\psi_g = \{\varphi_n\}^{N_g}_{n=1}$ with $\varphi_n = (t_{s,n}, t_{e,n})$. Here  $t_{s,n}$ and $t_{e,n}$ denote the start and end time index of the action instance $\varphi_n$ respectively.
The target of TAPG is to generate temporal proposals each covering a  ground truth action instance as accurate as possible.  Unlike the temporal action detection task, here the action instance categories need not be considered. 

Following previous TAPG methods~\cite{lin2019bmn,xu2020g,bai2020boundary}, we adopt a pre-trained encoding model to extract   features of the input video $\mathcal{X}$, which serve as inputs to our developed TAPG model.
In particular, 
for any given video $\mathcal{X}$, we split it into $T$ continuous and non-overlapped snippets.  
We then adopt a two-stream network~\cite{wang2016temporal} for extracting  $T$  snippet features of $C$-dimension, which are then  concatenated to form the video features $F \in \mathbb{R}^{T\times C}$ and fed into the our model for proposal generation.


\subsection{Augmented transformer with adaptive graph architecture.}
\label{ssec:transformer}
ATAG introduces a dual-path module, consisting of an augmented transformer and adaptive GCN for global and local temporal context capture, respectively.
The input feature $F$ is split into two parts along the channel dimension, denoted as $F^g$ and $F^l$, which are then fed into the two modules. 

\paragraph{Augmented transformer.}
We apply transformers to improve the semantic representation of snippet-level features by capturing long-range dependencies.
A vanilla transformer block is composed of a self-attention layer and position-wise feed-forward layers, performing multi-head attention to compute a weighted sum of input snippet features $F^g\in \mathbb{R}^{T\times C}$, resulting an augmented feature $\tilde{F}^g$ with global context. 
In particular, for each attention head, the attention map is computed by matching the transformed input features (a.k.a. the queries) $Q=f(F^g)$ to 
another transformation of the input features (a.k.a. the keys) $K=g(F^g)$, with $f$ and $g$ being learnable linear transformation. 
Namely,
\begin{equation}
\label{eqn_1}
    A = Attention(F) = \mathrm{softmax}(\frac{QK^\top }{\sqrt{d}}),
\end{equation}
where $A\in \mathbb{R}^{T\times T}$ is the generated attention map, and $d$ is the dimension for $Q$ and $K$.
The above attention map computation can find the relationship between features $Q$ and $K$ of global snippets and then aggregate their information into another linear transformation of input features $V=h(F^g)$ (a.k.a. the values).
The multi-head attention is followed by an FFN layer, which contains 
two  linear layers with ReLU activation and  a residual connection  after each layer. We also include layer normalization and dropout to facilitate training.

Directly employing the vanilla transformer presents two issues here. First, the main task of TAPG is the boundary regression, and the corresponding loss functions supervise proposal-level signal and cannot guide the transformer to effectively learn  long-term semantic relations for each snippet.
Second, according to  Eq.~\ref{eqn_1}, the computation of any element $A_{mn}$ in the attention map matrix $A$ depends on only the features of snippet m and n, namely, $F_m^g$ and $F_n^g$, which indicates that the attention map generation does not consider any temporal context, especially in limit transformer layers.
To solve the above issues, we add a snippet actionness loss and a front block to the vanilla transformer and it is dubbed augmented transformer.

The snippet actionness loss explicitly guides the transformer to learn effective long-range dependencies at snippet level. 
An actionness predictor is equipped upon the FFN, and it is used to predict the probability of an action instance existing in the input snippet, by minimizing the following binary classification loss $L_{a}$ over action/background categories: 
\begin{equation}
    L_{a}=\sum^T_{i=1}\alpha^+\cdot g_i^a\cdot \log p^a_i+\alpha^-\cdot(1-g_i^a)\cdot \log(1-p_i^a),
\end{equation}
where $\alpha^+={T}/\sum(g_i^a)$, $\alpha^-={T}/{\sum(1-g_i^a)}$, and $p^a_i$ and $g_i^a$ are the actionness predicted probability and action/background label of snippet $i$.

This loss function supervises the network to obtain snippet-wise classification only depending on each snippet feature from the output of transformer, which is critical to achieve noisy action instances localization. For those noisy snippets that have action labels, the network shall build correct relations among the noisy snippets and others to make correct decisions. Thus, this loss function helps the transformer learn how to selectively capture long-range dependencies and understand the complicated noisy action instances. 

The front block is a light-weight network consisting of three parts to expand the temporal receptive field. We first apply a gated linear unit~\cite{dauphin2017language}, followed by a parallel 1 $\times$ 1 convolutional layer and 3 $\times$ 1 average pooling layer with stride 1 to expand the receptive field, and the small 3 $\times$ 1 average pooling layer can also smooth the snippet-level feature to filter out the tiny noisy frames.
The last part is a convolutional layer with a large kernel size, for example, 7 $\times$ 1. To avoid over-fitting from large size kernels, we adopt the depth-wise separable convolution. We apply residual connection for each part and layer normalization behind each part. The structure of front block is depicted in the top-right corner in Fig.~\ref{figure:structure}.

\paragraph{Adaptive graph convolutional layer.}
To capture the local context, we design a new graph convolution layer to construct the local branch. We build a video graph $\mathcal{G}=\{\mathcal{V},\mathcal{E}\}$, where $\mathcal{V}=\{v_i\}_{i=1}^T$ and $\mathcal{E}$ represent the node and edge sets, respectively. Each node represents a snippet and each edge shows the dependency between two snippets. For local context modeling,  edges between two nodes are constructed  according to their temporal distance. The edge set is defined as:
\begin{equation}
    \mathcal{E}=\{(v_i, v_{i+k})|i \in \{1, ..., T\}; k \in  \{0, \pm{1},...,\pm{\delta}\}\},
\end{equation}
where $T$ is the number of snippets, and $\delta$ is defined as the maximum connection distance.
Our graph convolution operation $\mathcal{F}$ is defined as:
\begin{equation}
    \label{eq2}
    \mathcal{F}(F^l) = W\cdot F^l\cdot (A_a+A_d)\odot{M},
\end{equation}
where $W\in{\mathbb{R}^{C_{\mathrm{out}}\times C_{\mathrm{in}}}}$ includes trainable weights, $M$ is a fixed $T\times T$ binary mask matrix to limit the connection range according to   $\delta$, and $\odot$ is the dot product. 
Instead of using a pre-defined adjacency matrix, 
the matrix in Eq. \ref{eq2} is divided into two parts: $A_a$ and $A_d$ and treated differently.

The first part ($A_a$) is completely parameterized and optimized in the training phase. The elements in  $A_a$ can be arbitrary values without any constraints, which means that the edges of the graph are completely learned based on the training data and the positions of nodes in the video.


The second part ($A_d$) is a data-dependent graph which adaptively learns a unique graph for each video. Our GCN is used to capture local temporal context, therefore we pay more attention to the difference of features. To determine whether there is a connection between two nodes and how strong the connection is,  the difference of two nodes $m$ and $n$ is computed as: 
\begin{equation}
    A_{d, mn} = \frac{exp(\sigma({\theta}^T|f_m-f_n|))}{\sum_{i=1}^N exp(\sigma({\theta}^T|f_m-fi|))},
\end{equation}
where $\sigma$ is an activation function, ${\theta}^T$ is the trainable parameter vector to reduce the dimension of $f_m-f_n$ to $1$. The detailed visual description is shown in Figure~\ref{figure:structure}.


\paragraph{Global-local fusion.}
Early fusion and late fusion are widely used in two-stream feature fusion. In our method, we use the early fusion, \ie, the global and local features are fused by concatenation before the output module.

\subsection{Output module}
\label{ssec:output}
After obtaining global and local features, 
we feed them into our output module to generate the proposal boundaries and completeness scores, which are applied for proposal generation during inference.
\paragraph{Temporal boundary classification.}
Due to our global-local combination mechanism, the two-branch features 
have precise and discriminative action/background representation.
Here, we input them into two same convolutional networks consisting of two temporal convolutional layers and a sigmoid activation function to generate the probability of the start $p^s$ and end $p^e$ for each snippet, respectively. 

\paragraph{Completeness regression.} Besides the prediction of boundaries, the network is trained to predict the completeness of proposals to boost the final results. Our completeness regression network follows ~\cite{lin2019bmn} and also introduces the Boundary-Matching mechanism to generate completeness scores for densely distributed proposals. For each proposal, we sample $32$ features from the corresponding two-branch features. Sampled features are then input into the completeness prediction network to generate two completeness score maps $M^{cc}\in \mathbb{R}^{T\times D}$ for completeness classification, and $M^{cr}\in \mathbb{R}^{T\times D}$ for completeness regression, where $D$ is the maximum proposal duration and is set depending on the dataset.

\subsection{Training}
\label{ssec:training}


We apply weighted binary logistic regression loss function $L_{bl}$ for start and end classification losses, denoted as $L_s$ and $L_e$, where $L_{bl}$ is denoted as:
\begin{equation}
    \label{bl}
    L_{bl}=\sum^T_{i=1}(\alpha^+\cdot g_i\cdot \log p_i+\alpha^-\cdot(1-g_i)\cdot \log(1-p_i)),
\end{equation}
where $\alpha^+={T}/ \sum(g_i) $, $\alpha^-=  {T} / {\sum(1-g_i)}$, and $p_i$ and $g_i$ are the predicted probability and ground truth of $i$-th snippet. Following BMN~\cite{lin2019bmn}, our completeness loss $L_{com}$ consists of two different loss functions, binary classification loss and regression loss:
\begin{equation}
    L_{com} = L_{c}(G^c, M^{cc}) + \lambda L_r(G^c, M^{cr}),
\end{equation}
where $L_c$ is also a binary logistic regression loss function and its formula is as same as Eq. \ref{bl} and $L_r$ is $l_2$ loss, and $\lambda$ is set as $10$.

The model is trained in the form of a multi-task loss function, with the overall loss function defined as:
\begin{equation}
    \label{overall loss}
    L = L_{com} + \lambda_1 L_a + \lambda_2 L_s + \lambda_3 L_e,
\end{equation}
where $\lambda_1$, $\lambda_2$, and $\lambda_3$ are three scalars to balance the three terms. The method of label assignment will be described in detail in the supplementary material.

\subsection{Inference}
\label{ssec:inference}
During inference, a proposal set $\psi_p=\{\phi_n=(t_s,t_e,p^s_{t_s},p^e_{t_e},p^{cc}_{t_s,t_e},p^{cr}_{t_s,t_e})\}_{n=1}^N$ is generated, where $p^s_{t_s}$, $p^e_{t_e}$ are start and end probabilities in $t_s$ and $t_e$, and $p^{cc}_{t_s,t_e}$, $p^{cr}_{t_s,t_e}$ are completeness classification score and completeness regression score of proposal from $[t_s,t_e]$. To obtain final results, it is necessary to perform steps of score fusion and redundant proposal suppression.
\paragraph{Score fusion.}
In order to make full use of various predicted scores for each proposal $\phi_n$, we fuse its boundary probabilities and completeness scores by multiplication. The confidence score $p^f$ can be defined as:
\begin{equation}
    p^f=p^s_{t_s}\cdot p^e_{t_e}\cdot p^{cc}_{t_s,t_e}\cdot p^{cr}_{t_s,t_e}.
\end{equation}
Hence, the proposals set can be re-written as $\psi_p=\{\phi_n=(t_s,t_e,p^f)\}_{n=1}^N$.
\paragraph{Redundant proposals suppression.}
Because of our dense proposal generation, it is inevitable to output numerous redundant proposals which highly overlap with each other. Thus, based on the completeness score for proposals, we apply the Soft-NMS algorithm to remove redundant proposals. Candidate proposal set $\psi_p$ turns to be $\psi_p^\prime=\{\phi_n=(t_s,t_e,p^{f\prime})\}_{n=1}^{N^\prime}$, where $p^{f\prime}$ and $N^\prime$ is the final completeness score and the number of final proposals, respectively.

\section{Experiments}
\subsection{Datasets and setup}
\paragraph{Datasets.}
The evaluation is performed on two popular TAPG benchmark datasets, THUMOS14 \cite{jiang2014thumos} and ActivityNet1.3 \cite{Heilbron_2015_CVPR}.
We use the subset of THUMOS14 that provides frame-wise action annotations.  The model is trained with 200 untrimmed videos from its validation set and evaluated  using 213 untrimmed videos from the test set. This dataset is challenging due to the large variations of the frequency and duration of action instances across videos. 
ActivityNet1.3 is a large-scale dataset covering 200 complicated human activity classes with 19,994 untrimmed videos, which are divided into training, validation and test set in the ratio 2:1:1. The model is trained on the training set and evaluated with the validation set.

\paragraph{Implementation details.}
For feature encoding, following previous works~\cite{lin2019bmn,bai2020boundary}, we adopt two-stream network~\cite{wang2016temporal} pre-trained on training set of Activity1.3. The frame interval is set to $5$ and $16$ on THUMOS14 and Activity1.3, respectively. For ActivityNet1.3, we resize video feature sequences by linear interpolation to 100, \ie $T=100$, and set the maximum duration length as $100$. For THUMOS14, we slide the window on video feature sequence with $50\%$ overlap and $T=128$ and set the maximum duration length as $64$, which can cover $98\%$ action instances.

Adam~\cite{kingma2014adam} optimizer is adopted and the learning rate is set to $10^{-3}$ and decayed by a factor of 0.1 after every 10 epochs. To preserve the positional information in the transformer, we adopt sine positional encoding and add it to queries and keys. There are 8 heads in the multi-head attention module is $8$.  $\delta = 2$ in local branch. In Eq. \ref{overall loss}, the coefficients $\lambda_1$, $\lambda_2$ and $\lambda_3$ are all set to $1$. Early fusion is adopted unless otherwise specified.

\subsection{Temporal action proposal generation}
Temporal action proposal generation aims to produce high quality proposals that have high IoU with ground truth action instances and high recall. To verify proposal quality, Average Recall (AR) under multiple IoU thresholds are calculated. For ActivityNet1.3 and THUMOS14, the IoU thresholds used are $[0.5:0.05:0.95]$ and $[0.5:0.05:1.0]$, respectively. AR under different Average Number of proposals (AN) is defined as AR@AN, and the area under the AR vs. AN curve is named AUC. We use AR@AN and AUC as our metrics to evaluate our model.
\paragraph{Comparison with state-of-the-art methods.}
Our model is compared with state-of-the-art TAPG methods on THUMOS14 and ActivityNet1.3. Table~\ref{Table 1} shows the comparison on THUMOS14. To ensure a fair comparison, both C3D and two-stream features used by previous methods are adopted in our experiments. Following previous methods~\cite{lin2018bsn,lin2019bmn,lin2020fast,bai2020boundary}, we apply NMS and soft-NMS as  post-processing methods to evaluate our method, respectively. For both C3D and two-stream features, our results outperform other state-of-the-art methods by a large margin. Additionally, NMS achieves better average recall performance than soft-NMS under small proposal numbers.

The results on ActivityNet1.3 are summarized in Table~\ref{Table 2}. Our method again surpasses all of the leading methods. 

These experiments demonstrate the effectiveness of our model by adequately exploiting the global and local context.

\paragraph{Visualization and analysis.}
Fig.~\ref{visualization} visualizes some representative results on ActivityNet1.3 and THUMOS14 featuring different challenging cases. The proposals with the highest $k$  scores are visualized in each video, where $k$ is the number of ground truth. In the top video,  background frames and action frames have the similar scene and there are many irrelevant frames which are nearly the same as background frames in the action. However, our top-1 proposal still successfully aligns the position of ground truth action instance. 
Our network is trained to regard the noisy snippets in action instance as action. We concern that whether the network will overfit on the noisy frames and many background snippets are mistakenly treated as action instance. In the middle video, a cheering segment exists between the true action and background so it needs to be classified as background, which our model predicts correctly. Note that this is different from the case in  Fig.~\ref{figure1}, where the cheering segment is embedded in an action instance. The correct prediction made by our model shows that it correctly understands the semantic information and does not overfit on the noisy frames. The bottom video has multiple ground truth action instances, while our top-5 proposals perfectly cover them in an accurate way, suggesting the high quality of our generated proposals. More examples and comparison with the previous method are shown in the supplement material.

\begin{table}[t] 
\begin{center}
\small
\resizebox{!}{5.6cm}{
\begin{tabular}{c|c|p{1.3em}p{1.4em}p{1.4em}p{1.4em}p{1.6em}}
\toprule
\multirow{2}{2.5em}{Feature} & \multirow{2}{3em}{Methods} & \multicolumn{5}{c}{AR@AN} \\
& & @50 & @100 & @200 & @500 & @1000  \\
\midrule
\multirow{14}{2.5em}{C3D} 
& SCNN-prop \cite{shou2016temporal} & 17.22 & 26.17 & 37.01 & 51.57 & 58.20 \\
& SST \cite{Buch_2017_CVPR} & 19.90 & 28.36 & 37.90 & 51.58 & 60.27 \\
& TURN \cite{Gao_2017_ICCV} & 19.63 & 27.96 & 38.34 & 53.52 & 60.75 \\
& BSN \cite{lin2018bsn}+NMS & 27.19 & 35.38 & 43.61 & 53.77 & 59.50 \\
& BSN \cite{lin2018bsn}+SNMS & 29.58 & 37.38 & 45.55 & 54.67 & 59.48 \\
& MGG \cite{Liu_2019_CVPR} & 29.11 & 36.31 & 44.32 & 54.95 & 60.98  \\
& BMN \cite{lin2019bmn}+NMS & 29.04 & 37.72 & 46.79 & 56.07 & 60.96 \\
& BMN \cite{lin2019bmn}+SNMS & 32.73 & 40.68 & 47.86 & 56.42 & 60.44 \\
& DBG \cite{lin2020fast}+NMS & 32.55 & 41.07 & 48.83 & 57.58 & 59.55 \\
& DBG \cite{lin2020fast}+SNMS & 30.55 & 38.82 & 46.56 & 56.42 & 62.17 \\
& BC-GNN+NMS \cite{bai2020boundary} & 33.56 & 41.20 & 48.23 & 56.54 & 59.76 \\
& BC-GNN+SNMS \cite{bai2020boundary} & 33.31 & 40.93 & 48.15 & 56.62 & 60.41 \\
\cline{2-7}
& ATAG (Ours)+NMS & \bf{34.91} & \bf{42.34} & 48.64 & 57.96 & 62.03  \\
& ATAG (Ours)+SNMS & 34.47 & 41.92 & \bf{49.60} & \bf{58.49} & \bf{62.24}  \\
\midrule
\multirow{14}{2.5em}{2stream} 
& TAG \cite{zhao2017temporal} & 18.55 &29.00 & 39.61  & - & - \\
& TURN \cite{Gao_2017_ICCV} & 21.86 & 31.89 & 43.02 & 57.63 & 64.17  \\
& CTAP \cite{Gao_2018_ECCV} & 32.49 & 42.61 & 51.97 & - & -  \\
& BSN \cite{lin2018bsn}+NMS & 35.41 & 43.55 & 52.23 & 61.35 & 65.10 \\
& BSN \cite{lin2018bsn}+SNMS & 37.46 & 46.06 & 53.21 & 60.64 & 64.52 \\
& BMN \cite{lin2019bmn}+NMS & 37.15 & 46.75 & 54.84 & 62.19 & 65.22  \\
& BMN \cite{lin2019bmn}+SNMS & 39.36 & 47.72 & 54.70 & 62.07 & 65.49  \\
& MGG \cite{Liu_2019_CVPR} & 39.93 & 47.75 & 54.65 & 61.36 & 64.06 \\
& DBG \cite{lin2020fast}+NMS & 40.89 & 49.24 & 55.76 & 61.43 & 61.95 \\
& DBG \cite{lin2020fast}+SNMS & 37.32 & 46.67 & 54.50 & 62.21 & 66.40 \\
& BC-GNN+NMS \cite{bai2020boundary} & 41.15 & 50.35 & 56.23 & 61.45 & 66.00 \\
& BC-GNN+SNMS \cite{bai2020boundary} & 40.50 & 49.60 & 56.33 & 62.80 & 66.57 \\
\cline{2-7}
& ATAG (Ours)+NMS & \bf{43.60} & \bf{52.21} & \bf{59.67} & 65.98 & 69.24  \\
& ATAG (Ours)+SNMS & 43.52 & 51.86 & 59.48 & \bf{66.04} & \bf{70.28}  \\
\bottomrule
\end{tabular}
}
\end{center}
\vspace{-4mm}
\caption{Comparison of our model with state-of-the-art methods on THUMOS14 testing set. All the results are reported in percentage. SNMS stands for Soft-NMS. }
\label{Table 1}
\end{table}

\begin{table} [t]
\small
\begin{center}
\small
\resizebox{!}{0.95cm}{
\begin{tabular}{c|cccccc}
\toprule
Method &\cite{lin2018bsn} & \cite{lin2019bmn} & \cite{Liu_2019_CVPR} & \cite{bai2020boundary} & \cite{zhao2020bottom} & ours \\
\midrule
AR@100(val) & 74.16 & 75.01 & 74.54 & 76.73 & 75.27 & \bf{76.75} \\
AUC(val) & 66.17 & 67.10 & 66.43 & 68.05 & 66.51  & \bf{68.50} \\
AUC(test) & 66.26 & 67.19 & 66.47 & - & - & \bf{68.45} \\

\bottomrule
\end{tabular}
}
\end{center}
\vspace{-5mm}
\caption{Comparison of our model with state-of-the-art methods BSN~\cite{lin2018bsn}, BMN~\cite{lin2019bmn}, MGG~\cite{Liu_2019_CVPR}, BC-GCN~\cite{bai2020boundary}, BUMR~\cite{zhao2020bottom} on ActivityNet1.3. All the results are reported in percentage.}
\label{Table 2}
\end{table}

\begin{figure}[t]
\includegraphics[width=1.0\linewidth]{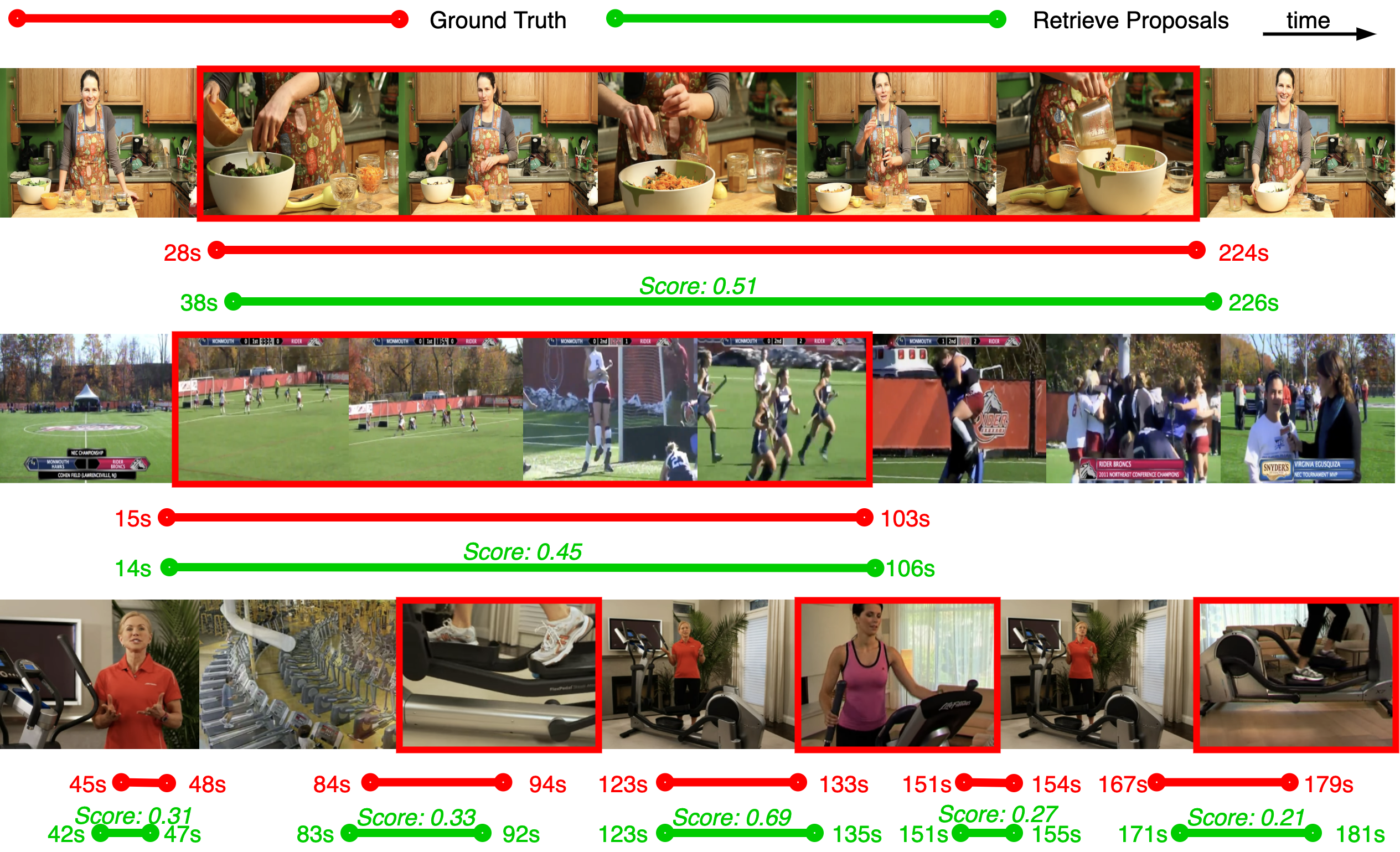}
\centering
\caption{Visualization examples of generated proposals on ActivityNet1.3 and THUMOS14. The red boxes highlight some frames in the action instances. }
\label{visualization}
\vspace{-4mm}
\end{figure}

\subsection{Ablation study}
To verify the effectiveness of our method, we conduct the following ablation studies on the validation set of ActivityNet1.3 dataset. 

\paragraph{Component analysis.}
The augmented transformer and adaptive GCN are the core components in our method. We evaluate several variants of this model by ablating its augmented transformer and GCN in Table~\ref{Ablation1}. \textit{Base} represents the version without augmented transformer and adaptive GCN. We can see that any component can individually improve the performance. The existence of snippet actionness loss and front block can boost the effect of vanilla transformer, which demonstrates that they can assist the transformer to learn long-range dependencies. For local context capture, the result shows that adaptive graph is beneficial for this task and removing any adjacency matrix will degrade the performance. With the global and local fusion, the full model reaches the best result.

\paragraph{Early fusion \vs. late fusion.}
We explore other fusion strategies. Specifically, in the early fusion, besides our early fusion with concatenation, we also evaluate the global and local features are fused by summation; in the late fusion, each path is individually fed into independent prediction networks of the output module and concatenate the dual-branch features till entering classifiers to aggregates predictions of boundaries and completeness scores.
We show the results in Table~\ref{Ablation1}. The Early fusion with concatenation outperforms other fusion strategies.

\paragraph{Number of transformer layers.}
Stacking multi-layer transformers normally brings better performance, which is demonstrated in other tasks, for instance, NLP~\cite{vaswani2017attention,devlin2018bert} and object detection~\cite{carion2020endtoend}. We stack different numbers of transformers and show their results in Table~\ref{Ablation2}. Different from other tasks, more transformers in our model do not improve the performance. We believe the reason is that the scale of datasets in this task is too small to train a deeper transformer network. Although there are about 20k videos in ActivityNet1.3, the density of action instances is low and each video includes only 1.5 action instances on average, which is far less than the datasets in other tasks.

\paragraph{Our GCN \vs. other schemes.}
For the local context capture, we use a GCN with a hybrid adaptive adjacency matrix. Further, we experiment with different structures to replace our GCN, including (i) a general GCN with predefined adjacency matrix, (ii) a self-attention style GCN where the adjacency matrix $A$ is calculated by $A=softmax(X^TW_\theta^TW_\phi X)$ where $X$ is the input feature sequence and $W_\theta$ and $W_\phi$ are trainable weights, and (iii) a convolutional layer. Table~\ref{Ablation3} compares all these variants, with our choice outperforming other two variations.

\begin{table} [t]
\begin{center}
\small
\begin{tabular}{c|cc}
\toprule
Variants of model & AR@100 & AUC  \\
\midrule
Base & 75.10 & 66.80  \\
Base + VT & 75.42 & 67.48 \\
Base + VT + Snippet actionness loss & 75.76 & 67.74 \\
Base + VT + Front block & 75.85 & 67.78 \\
Base + Augmented transformer & 76.19 & 68.01 \\
Base + GCN wo/$A_a$ & 75.50 & 67.31\\
Base + GCN wo/$A_d$ & 75.58 & 67.54\\
Base + GCN & 75.79 & 67.81 \\
\midrule
Late fusion & 76.26 & 68.25\\
Early fusion - summation & 75.99 & 67.88  \\
Early fusion - concatenation & 76.75 & 68.50 \\
\bottomrule
\end{tabular}
\end{center}
\vspace{-5mm}
\caption{Performance evaluation on different components of our model and comparison of different fusion strategies. VT is short for vanilla transformer.}
\label{Ablation1}
\end{table}

\begin{table} [t]
\begin{center}
\small
\begin{tabular}{c|cc}
\toprule
No. of Transformer & AR@100 & AUC  \\
\midrule
1 & 76.75 & 68.50  \\
2 & 76.64 & 68.35 \\
3 & 75.94 & 67.83 \\

\bottomrule
\end{tabular}
\end{center}
\vspace{-5mm}
\caption{Performance evaluation on different number  of  transformer  layers.}
\label{Ablation2}
\end{table}

\begin{table} [t]
\small
\begin{center}
\begin{tabular}{c|cc}
\toprule
 Structure & AR@100 & AUC  \\
\midrule
General GCN & 75.28 & 67.24  \\
Self-attention GCN & 74.76 & 66.50 \\
Convolutional layer & 75.20 & 67.20 \\
Our GCN & 75.79 & 67.81 \\
\bottomrule
\end{tabular}
\end{center}
\vspace{-5mm}
\caption{Comparison of our GCN with other schemes.}
\label{Ablation3}
\end{table}


\subsection{Temporal action detection with our proposals}
The important usage of action proposals is for temporal action detection. Thus, evaluating the performance of proposals in the temporal action detection task is another aspect to verify the quality of proposals. We adopt the evaluation metric Mean Average Precision (mAP) from the temporal action detection task. mAP with IoU thresholds $[0.5:0.05:0.95]$ are used on ActivityNet1.3 and mAP with IoU thresholds $\{0.3,0.4,0.5,0.6,0.7\}$ are used on THUMOS14.

We adopt the two-stage ``detection by classifying proposals'' temporal action detection framework. On ActivityNet1.3, the proposals are ranked according to their confidence scores and top 100 proposals per video are selected. Then, for each video, its top-1 video-level classification result will be obtained from ~\cite{zhao2017cuhk} and all the proposals of this video share this classification result as their action classes. On THUMOS14, we adopt top-2 video-level classification results generated by UntrimmedNet~\cite{Wang_2017_CVPR} as labels for top 200 proposals per video.

The results on ActivityNet1.3 and THUMOS14 are shown in Table~\ref{Table Activity} and Table~\ref{Table thumos}, respectively. Our method achieves new state-of-the-art results on both datasets, which validates the quality of our proposals and the effectiveness of our method. The high mAP reflects that our model can predict the good proposals with high scores, and reduce the number of false positives. 
\begin{table} [t]
\begin{center}
\begin{tabular}{c|cccc}
\toprule
Method & 0.5 & 0.75 & 0.95 & Average \\
\midrule
CDC~\cite{Shou_2017_CVPR} & 43.83 & 25.88 & 0.21 & 22.77  \\
SSN~\cite{xiong2017pursuit} & 39.12 & 23.48 & 5.49 & 23.98  \\
BSN~\cite{lin2018bsn}+\cite{zhao2017cuhk} & 46.45 & 29.96 & 8.02 & 30.03 \\
BMN~\cite{lin2019bmn}+\cite{zhao2017cuhk} & 50.07 & 34.78 & 8.29 & 33.85 \\
GTAD~\cite{xu2020g}+\cite{zhao2017cuhk} & 50.36 & 34.60 & 9.02 & 34.09 \\
BC-GNN~\cite{bai2020boundary}+\cite{zhao2017cuhk} & 50.56 & 34.75 & 9.37 & 34.26 \\
\midrule
ATAG (Ours)+\cite{zhao2017cuhk} & \bf{50.92} & 
\bf{35.35} & \bf{9.71} & \bf{34.68}\\
\bottomrule
\end{tabular}
\end{center}
\vspace{-5mm}
\caption{Action detection results on validation set of ActivityNet1.3 dataset in terms of mAP at IoU 0.5, 0.75 and 0.95, and average mAP.}
\label{Table Activity}
\end{table}

\begin{table} [t]
\begin{center}
\begin{tabular}{c|ccccc}
\toprule
Method & 0.7 & 0.6 & 0.5 & 0.4 & 0.3 \\
\midrule
TURN~\cite{Gao_2017_ICCV} & 6.3 & 14.1 & 24.5 & 35.3 & 46.3  \\
BSN~\cite{lin2018bsn} & 20.0 & 28.4 & 36.9 & 45.0 & 53.5 \\
MGG~\cite{Liu_2019_CVPR} & 21.3 & 29.5 & 37.4 & 46.8 & 53.9 \\
BMN~\cite{lin2019bmn} & 20.5 & 29.7 & 38.8 & 47.4 & 56.0 \\
GTAD~\cite{xu2020g} & 23.4 & 30.8 & 40.2 & 47.6 & 54.5 \\
BC-GNN~\cite{bai2020boundary} & 23.1 & 31.2 & 40.4 & 49.1 & 57.1 \\
\midrule
ATAG (Ours) & \bf{28.0} & \bf{38.0} & \bf{47.3} & \bf{53.1} & \bf{62.0}\\
\bottomrule
\end{tabular}
\end{center}
\vspace{-5mm}
\caption{Comparison between our approach and other temporal action detection methods on THUMOS-14.}
\label{Table thumos}
\end{table}

\section{Conclusion}
In this paper, we propose an augmented transformer with adaptive graph network (ATAG) for temporal action proposal generation, which presents an augmented transformer and a new adaptive GCN to capture long-range temporal context and local temporal context, respectively. Our augmented transformer can effectively enhance the video understanding and noisy action instance localization. For local context capture, the newly designed adaptive GCN with two types of matrices is leveraged to build local temporal relationships. Extensive experiments show that our model achieves new state-of-the-art performance in temporal action proposal generation and action detection on THUMOS14 and ActivityNet1.3 datasets. 
\section*{Acknowledgement}
This work was supported by Alibaba Group through Alibaba Research Intern Program.

{\small

\bibliographystyle{IEEEtran}
}

\clearpage
\section{Supplementary }

\subsection{Label assignment}
The methods of label assignment in each loss functions are described in detail below. 

For boundary classification loss, we need to generate ground truth of boundary label including start $g^s$ and end $g^e$. Given a ground truth action instance $\phi_n=(t_s, t_e)$, we denote its start and end region as $r_s=[t_s-1.5\Delta t, t_s+1.5\Delta t]$ and $r_e=[t_e-1.5\Delta t, t_e+1.5\Delta t]$ respectively, where $\Delta t$ is the temporal interval between two adjacent snippets. Then, we compute overlap ratio IoR of each snippet interval with $r_s$ and $r_e$ separately, where IoR is
defined as the overlap ratio with ground truth proportional. If one snippet interval is overlapped with multiple actions, we take the maximum IoR. Finally, we assign positive labels to the locations with IoR $>0.5$; otherwise negative labels.

For actionness classification loss, we need to generate ground truth of actionness label $g^a$. Similar to boundary label generation, we calculate the IoR of each snippet interval with ground truth of action instances and use the same threshold $0.5$ to assign a positive or negative label.

For proposal completeness loss, we need to generate proposal completeness label map $G^c$. For a proposal $\phi_{i,j}=(t_j, t_j+t_i)$, we compute its Intersection-over-Union (IoU) with all the ground truth action instances, and denote the maximum IoU as $G^c[i,j]$.
\subsection{Additional visualization}

Figure \ref{visualization} illustrates more visualization examples and comparison with classical method BMN [\textcolor{green}{16}]. The proposals of our method and BMN\footnote{The reproduction code we use is from \url{https://github.com/JJBOY/BMN-Boundary-Matching-Network}, whose performance is sightly higher than paper's.} with the highest $k$  scores are visualized in each video, where $k$ is the number of ground truth.  All the examples belong to complicated scenarios where interested actions involve irrelevant frames and background clutters. Traditional BMN fails to be robust for noisy frames and outputs incomplete proposal in the first four videos. In the last two videos, although its predictions cover the whole action instances, we speculate the reason is that the network overfits the noisy frames since those proposals also contain massive background frames. By comparison, our method presents precise boundary predictions in all the cases. Those examples adequately demonstrate that our method can understand video semantics and deal with complicated and noisy action instances by modeling long-range and local temporal context, by equipping our novel augmented transformer and adaptive GCN; in contrast, previous methods, such as BMN, focusing on the single local temporal context lack this capacity.

\subsection{Detailed architecture}
Table \ref{architecture} presents the architecture of our actionness predictor in augmented transformer and output module including completeness prediction network and boundary prediction network.


\begin{table} [t]
\begin{center}
\small
\resizebox{!}{3.5cm}{
\begin{tabular}{c|ccp{1.5em}p{1.5em}c}
\toprule
layer & kernel size & stride & group & act & output size \\
\midrule
\multicolumn{6}{c}{Actionness Predictor} \\
\midrule
conv1d & 3 & 1 & 4 & ReLU & $256{\times}T$ \\
\midrule
conv1d & 1 & 1 & 1 & Sigmoid & $1{\times}T$\\
\midrule
\multicolumn{6}{c}{Completeness Prediction Network} \\
\midrule
BM layer & - & - & - & - & $256{\times}32{\times}D{\times}T$\\
conv3d & (32,1,1) & (32,0,0) & 1 & ReLU & $512{\times}1{\times}D{\times}T$ \\
squeeze & - & - & - & - & $512{\times}D{\times}T$ \\
conv2d & (1,1) & (1,1) & 1 & ReLU & $128{\times}D{\times}T$ \\
conv2d & (3,3) & (1,1) & 1 & ReLU & $128{\times}D{\times}T$ \\
conv2d & (3,3) & (1,1) & 1 & ReLU & $128{\times}D{\times}T$ \\
conv2d & (1,1) & (1,1) & 1 & Sigmoid & $2{\times}D{\times}T$\\
\midrule
\multicolumn{6}{c}{Boundary Prediction Network} \\
\midrule
conv1d & 3 & 1 & 4 & ReLU & $256{\times}T$\\
\midrule
conv1d & 1 & 1 & 1 & Sigmoid & $2{\times}T$\\

\bottomrule
\end{tabular}
}
\end{center}
\vspace{-5mm}
\caption{The detailed architectures of some modules in our method.$T$ is the number of snippets and $D$ represents the maximum duratin length.}
\label{architecture}
\end{table}

\begin{figure*}[t]
\includegraphics[width=1.0\linewidth]{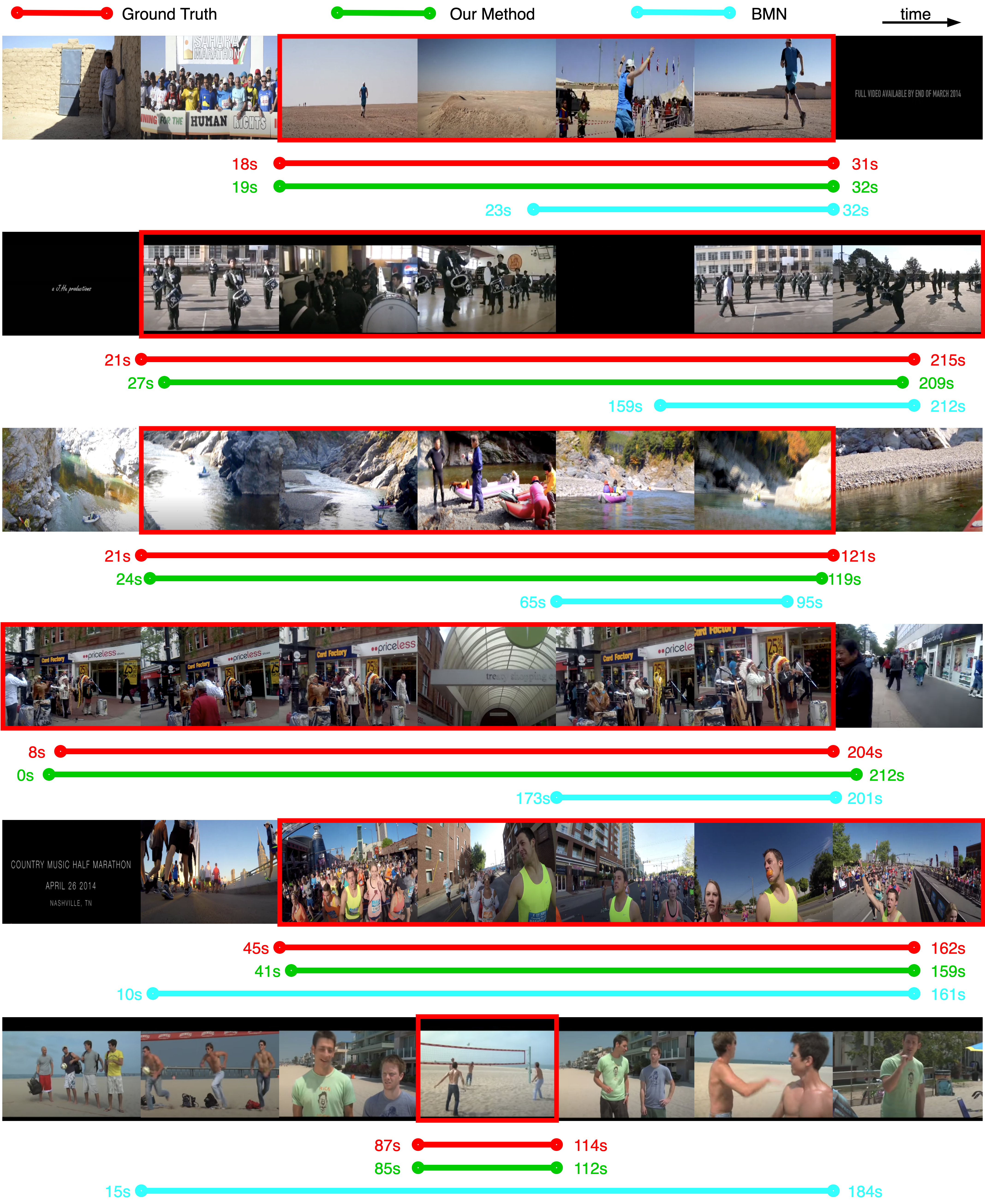}
\centering
\caption{Visualization examples of generated proposals on ActivityNet1.3 and THUMOS14. The red boxes highlight some frames in the action instances. }
\label{visualization}
\end{figure*}

\end{document}